\documentclass[letterpaper, 10 pt, conference]{ieeeconf}  

\IEEEoverridecommandlockouts                              
\let\labelindent\relax         

\usepackage{times}
\usepackage{graphicx}
\usepackage{epstopdf}

\usepackage{mdwmath}
\usepackage{mdwtab}
\usepackage{rotating}
\usepackage{caption}
\usepackage{subcaption}

\usepackage{amssymb}
\usepackage{amsfonts}
\usepackage{amsmath}
\usepackage{bm}

\usepackage{algorithm}
\usepackage{cite}
\usepackage{algorithmic}

\usepackage{multirow} 
\usepackage{multicol}
\usepackage{array}

\usepackage{standalone}
\usepackage{booktabs} 
\usepackage{float}

\usepackage{siunitx} 

\usepackage{accents}

\usepackage{pgfplotstable} 
\usepackage{verbatim}
\usepackage{isomath}
\usepackage{overpic}
\usepackage{tikz}
\usetikzlibrary{shapes,calc,patterns,
	decorations.pathmorphing,
	decorations.markings}

\tikzstyle{spring}=[very thick,decorate,decoration={zigzag,pre length=2,post
	length=2,segment length=6}]

\tikzstyle{damper}=[thick,decoration={markings, 
	mark connection node=dmp,
	mark=at position 0.5 with 
	{
		\node (dmp) [thick,inner sep=0pt,transform shape,rotate=-90,minimum
		width=15pt,minimum height=3pt,draw=none] {};
		\draw [thick] ( $(dmp.north east)+(2pt,0)$ ) -- (dmp.south east) -- (dmp.south
		west) -- ( $(dmp.north west)+(2pt,0)$ );
		\draw [thick] ( $(dmp.north)+(0,-5pt)$ ) -- ( $(dmp.north)+(0,5pt)$ );
	}
}, decorate]

\makeatletter\newcommand{\manuallabel}[2]{\def\@currentlabel{#2}\label{#1}}\makeatother

\usepackage{color}
\usepackage{xcolor}

\colorlet   {lightorange}{orange!20}
\colorlet   {lightgrey}  {gray!20}

\usepackage{amssymb}
\usepackage{mathtools}

\mathchardef\mhyphen="2D   

\newcommand{\RNum}[1]{\uppercase\expandafter{\romannumeral #1\relax}}

\graphicspath{
{figures/}
}

\usepackage[inline]{enumitem}
\setlist{nolistsep}
\newcommand{\il}[1]{\begin{enumerate*}[label=(\roman*)]#1\end{enumerate*}}

\usepackage[normalem]{ulem}                                        
\usepackage{marginnote}
\usepackage[textwidth=10ex,colorinlistoftodos]{todonotes}

\colorlet{fwu}{red}
\colorlet{ywu}{blue}
\colorlet{zbing}{green}
\colorlet{xchen}{pink}
\colorlet{hs}{magenta}

\title{\LARGE \bf%
SharedAssembly: A Data Collection Approach via Shared Tele-Assembly 
}

\author{Yansong Wu$^{1,\dagger}$, Xiao Chen$^{1,\dagger}$, Yu Chen$^1$, Hamid Sadeghian$^{1,*}$, Fan Wu$^{1,3,*}$, Zhenshan Bing$^{1,4}$, \\ Sami Haddadin$^2$,  Alois Knoll$^1$ 
\thanks{$^{\dagger}$ The authors contribute equally.}
\thanks{$^*$ Corresponding author: Fan Wu, Hamid Sadeghian}
\thanks{$^1$ Technical University of Munich}
\thanks{$^2$ Mohamed bin Zayed University of Artificial Intelligence}
\thanks{$^3$ Shanghai University}
\thanks{$^4$ Nanjing University}
}

\begin{document}
\maketitle

\thispagestyle{empty}
\pagestyle{empty}
\begin{abstract}

High-precision, tight-clearance assembly demonstrations are indispensable for training tactile-aware robotic foundation models, yet their acquisition is heavily bottlenecked by the high operational barriers of conventional teleoperation. To bridge this gap, we propose SharedAssembly, a novel shared-autonomy bilateral teleoperation framework that embeds assembly-specific intelligence across both leader and follower sides to facilitate scalable data collection. Rigorous real-world user studies on challenging sub-millimeter tasks show that SharedAssembly achieves an exceptional 97\% assembly success rate while significantly boosting completion efficiency. Notably, these performance gains become even more pronounced as the assembly clearance shrinks. Furthermore, our framework effectively eliminates the expertise gap, enabling novice operators to match or even outperform expert operators using conventional systems. By minimizing the skill barrier, SharedAssembly provides an efficient, robust, and accessible solution for large-scale data harvesting in contact-rich manipulation.
\end{abstract}
\section{Introduction}

~\ref{tab: datasets}. However, teleoperation performance on contact-rich tight-clearance assembly tasks, which play a critical role in both service and industry robotics~\cite{whitney2004mechanical}, remains notoriously challenging due to the stringent requirements on motion precision~\cite{inoue2017deep} and force regulation~\cite{wu2025tacdiffusion}. 

Thanks to the pioneering open-source robot learning datasets~\cite{brohan2022rt, sharma2018multiple, jang2022bc, fang2023rh20t, bharadhwaj2024roboagent, walke2023bridgedata, khazatsky2024droid, o2024open}, Vision-Language-Action (VLA) models have achieved remarkable success across a wide range of general-purpose manipulation tasks~\cite{zitkovich2023rt,kim2024openvla,intelligence2025pi_}. However, the majority of existing robot learning datasets are dominated by vision-centric manipulation tasks. As a result, current VLA models still struggle with contact-rich manipulation tasks, particularly tight-clearance assembly~\cite{sliwowski2025reassemble}. This limitation has motivated researchers to explore the integration of tactile information into VLA frameworks, enabling robots to reason about physical interactions beyond visual observations. Since the introduction of ForceVLA~\cite{yu2025forcevla} in 2025, tactile-enhanced Vision-Language-Action models have rapidly emerged as an active research direction. In a remarkably short span, numerous frameworks have been proposed by different research groups~\cite{yu2025forcevla,li2026forcevla2,li2026favla,zhang2026craft,zhao2026fd,huang2025tactile,zhang2025vtla,zhang2025ta,cheng2025omnivtla,zhang2026tacvla,wang2026tacmamba,bi2025vla,gubernatorov2026hapticvla,zhang2026compliantvla}, highlighting the growing recognition that tactile information is essential for robust contact-rich manipulation. For clarity, we refer to this family of tactile-enhanced robotic foundation models as Vision-Tactile-Language-Action (VTLA) models throughout this paper.

This paradigm shift, however, exposes a critical dilemma in data acquisition: Next-generation VTLA models demand high-quality tight-clearance demonstrations, yet collecting such data remains notoriously difficult for human operators. While loose-clearance tasks are relatively easy to teleoperate, policies trained on such data typically exhibit severe performance degradation when deployed in tight-clearance scenarios due to the strict requirements on positioning accuracy, contact-state perception, and force regulation. Conversely, demonstrations captured directly from high-precision, tight-clearance environments yield inherently richer interaction features that are indispensable for training generalizable policies. However, executing these contact-rich, sub-millimeter tasks via conventional bilateral teleoperation places an immense cognitive and physical burden on human operators. This operational bottleneck directly leads to low execution success rates and poor efficiency in data acquisition, resulting in a severe scarcity of high-quality, tight-clearance assembly datasets in the robot learning community~\cite{sliwowski2025reassemble}. 


\begin{table} [b]
\setlength{\tabcolsep}{1pt}
\centering
\begin{tabular}{lccccl}
\toprule
\textbf{Dataset} & \# \textbf{Traj}. & \# \textbf{Skills} & \textbf{Collection Method}\\ 
\midrule
RT1~\cite{brohan2022rt}    &     130k      &    8     &        Human Teleoperation    \\
MIME~\cite{sharma2018multiple}      &      8.3k     &    20              &     Kinesthetic Teaching      \\ 
BC-Z~\cite{jang2022bc} & 26k &  3 &  Human Teleoperation \\
RH20T~\cite{fang2023rh20t}  &     110k&         42&                                   Human Teleoperation\\
RoboSet~\cite{bharadhwaj2024roboagent}       &   98.5k        &     12           &     30\% Human / 70\% Scripted       \\
BridgeData V2~\cite{walke2023bridgedata}&     60.1k      & 13          &     85\% Human / 15\% Scripted       \\
DROID~\cite{khazatsky2024droid}      &     76k      &    86    &       Human Teleoperation         \\ 
\textcolor{gray}{Open X-Embodiment~\cite{o2024open}$^{\dagger}$}   &     \textcolor{gray}{1.4M}    &    \textcolor{gray}{217}   &  \textcolor{gray}{Dataset Aggregation} \\
\bottomrule
\end{tabular}
\caption{Summary of existing real-world datasets for training robot model. The notation ``\#" refers to the number. \textcolor{gray}{$^{\dagger}$only the aggregation of existing datasets instead of an independent novel dataset.}}
\label{tab: datasets}
\end{table}




To address this challenge, we propose \textit{SharedAssembly}, a novel shared-autonomy bilateral teleoperation framework designed for tight-clearance assembly and high-quality demonstration collection. By embedding assembly-specific intelligence into both the leader and follower sides of the system, our method provides seamless autonomous assistance during tele-assembly. Extensive user studies on real-world, sub-millimeter-clearance tasks demonstrate that \textit{SharedAssembly} significantly improves both assembly success rates and data collection efficiency across operators of varying experience levels. Notably, these benefits become increasingly pronounced as the assembly clearance decreases. By drastically reducing the skill barrier inherent in teleoperated assembly, our framework effectively democratizes high-precision data collection, enabling novice operators to consistently achieve expert-level performance. This collective advancement provides a robust and accessible solution for large-scale data harvesting in contact-rich manipulation tasks. 

The main contributions of this paper are:
\begin{itemize}
    \item Design and implementation of \textit{SharedAssembly}, a novel shared-autonomy bilateral teleoperation framework that embeds assembly-specific intelligence to facilitate large-scale, high-quality data collection for contact-rich tasks. 
    \item Formulation of a force-domain guidance mechanism that seamlessly blends human inputs with autonomous force regulation, significantly boosting teleoperation success rates and efficiency. 
    \item A comprehensive real-world user study involving operators of varying expertise, demonstrating the critical role of force feedback in enhancing task performance, furthermore validating that the proposed framework effectively democratizes high-precision tele-assembly by enabling novices to achieve expert-level efficiency. 
\end{itemize}


\section{Related Works}
In this section, we focus our review on \il{\item Teleoperation System for Data Collection, and \item Shared Teleoperation}.

\subsection{Teleoperation System for Data Collection \label{sec: Teleoperation}} 

As shown in Table~\ref{tab: datasets}, human teleoperation has emerged as the predominant method for acquiring real-world robotic datasets at scale. Based on the direction of the robot signal flow, the teleoperation system can be categorized into two types: 1) \textbf{Unilateral Teleoperation} in which the robot signal is transmitted solely from the leader to the follower, and 2) \textbf{Bilateral Teleoperation}, which includes a feedback loop from the follower to the leader, enabling the operator to perceive the remote environment.



The majority of the teleoperation systems utilized in data collection employ input devices such as space mouse or virtual reality (VR) remotes \cite{brohan2022rt, jang2022bc, bharadhwaj2024roboagent, walke2023bridgedata, khazatsky2024droid}. However, these devices lack force feedback, which complicates the execution of contact-rich tasks and increases the risk of unsafe interactions. Furthermore, due to workspace mismatches, the follower robot is prone to encountering singularities. To mitigate these issues, some studies have proposed input devices designed based on the kinematics of the follower robot \cite{aldaco2024aloha, zhao2024aloha, wu2024gellogenerallowcostintuitive} which also lack force feedback. Bilateral teleoperation, by incorporating force feedback, enhances user interaction with the remote environment, thereby improving task performance and increasing demonstration success rates \cite{nitsch2012meta}. Recognizing the importance of force feedback, recent research has incorporated it into teleoperation devices  \cite{sujit2025improving}. 

Haptic devices are commonly employed in teleoperation systems when force feedback is required \cite{fang2023rh20t, sliwowski2025reassemble}, facilitating data collection for assembly tasks. However, due to morphological differences between haptic devices and execution robots, as well as the unrealistic interaction forces they generate, demonstrations using these devices often lack intuitiveness \cite{pervez2017novel}. To address these limitations, this study adopts a teleoperation system consisting of two identical robot arms, following the approach in \cite{chen2024online}, to ensure maximum intuitiveness and transparency, while enabling accurate measurement of the interaction forces of follower robot.





\subsection{Shared Teleoperation}

The direct teleoperation of multi \textit{Degrees-of-Freedom (DoFs)} tasks is challenging, even in a system with high transparency. It often requires extensive training and demands significant mental acuity and proficiency. To address this, shared autonomy leverages the prior knowledge to assist human operators from different perspectives \cite{selvaggio2021autonomy}. This approach has been applied to the teleoperation system to improve safety \cite{Marinho2019VirtualFA} and increase intuitiveness \cite{laghi2018shared}. 
In \cite{quere2020shared} a shared control template is proposed where some task space coordinates are fixed or constrained to assist the users in task execution. Additionally, shared autonomy can mitigate the influence of time delays and improve the task performance \cite{chen2024network}, leading to stability and ensuring that the delayed visual feedback does not degrade the performance. 
In this work, we follow \cite{selvaggio2021autonomy}, where shared control employs a fixed level of autonomy, whereas shared autonomy adapts the autonomy level according to the task and interaction state.





\section {Preliminaries: Bilateral Teleoperation}
In this section, the basic position-force bilateral teleoperation architecture is first presented. The proposed shared bilateral teleoperation is then introduced in the next section. 

\begin{figure}[t]
    \centering
    \includegraphics[width=0.99\linewidth]{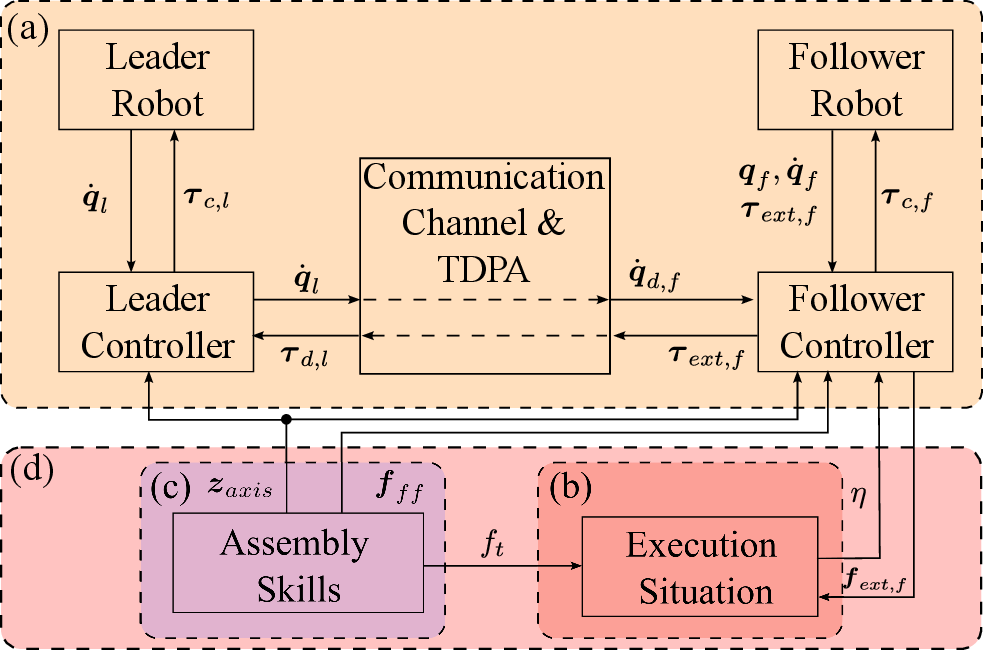}
    \caption{Overview of the SharedAssembly framework. (a) Bilateral teleoperation architecture. (b) Task execution state understanding module. (c) Assembly skill library. (d) Autonomy allocation module. } 
    \label{fig:teleop}
\end{figure}

A basic Position-Force (P-F) architecture for a bilateral teleoperation system is illustrated in block (a) of Fig. \ref{fig:teleop}. The joint velocity of the leader robot is transmitted through the communication channel to the {follower robot} as its desired velocity to track. In the other direction, the interaction force between the {follower robot} and the environment is transmitted back to the {leader robot} and applied to the operator's hand to feel the remote environment. 

Consider a torque-controlled robot with $n$-DoFs, and ignoring the joint friction, the dynamics of a robot manipulator in joint space is written as:
\begin{equation}
    \bm{M}_{i}(\bm{q}_i) \Ddot{\bm{q}}_i + \bm{c}_i(\bm{q}_i, \Dot{\bm{q}}_i) + \bm{g}_i(\bm{q}_i) = \bm{\tau}_{c,i} + \bm{\tau}_{ext,i} 
\end{equation}
where $\bm{M}_i \in \mathbb{R}^{n \times n}$ corresponds to the mass matrix, $\bm{c}_i \in \mathbb{R}^{n}$ indicates the Coriolis/centrifugal vector. The vectors $\bm{q}_i, \Dot{\bm{q}}_i, \Ddot{\bm{q}}_i \in \mathbb{R}^{n}$ represent the robot joint position, velocity, and acceleration. The parameter $\bm{\tau}_{c,i} \in \mathbb{R}^{n}$ is the command torque to the robot and $\bm{\tau}_{ext,i} \in \mathbb{R}^{n}$ is the external torque applied on the robot. The subscript $i\in \{l, f\}$  represents \textit{leader} and \textit{follower}, respectively.

\subsubsection{Leader}

As the leader robot serves as an intuitive input device, it is expected to provide the operator with a lightweight and responsive interface while simultaneously conveying force feedback from the follower. The control on the leader side is mainly a gravity compensation controller given by,
\begin{equation}
\begin{aligned}
    \bm{\tau}_{c,l} &= \bm{g}_l(\bm{q}_l) + \bm{c}_l(\bm{q}_l,  \dot{\bm{q}}_l) + \bm{\tau}_{d,f} \label{B_leader},\\
    \bm{\tau}_{d,f} &:= \bm{\tau}_{ext,f},
\end{aligned}
\end{equation} 
where $\bm{\tau}_{d,f}$ denotes the torque generated by external forces on the follower's joints transmitted back to the leader.

\subsubsection{Follower}
The follower robot is designed to synchronize with the leader's movements. A joint impedance control strategy is employed to track the leader's position and velocity, as follows:
\begin{equation}
    \bm{\tau}_{c,f} = \bm{K}_q \bm{e}_q + \bm{D}_q \dot{\bm{e}}_q + \bm{c}_f(\bm{q}_f, \dot{\bm{q}}_f) + \bm{g}_f(\bm{q}_f), 
\end{equation}
where the joint error and velocity error are expressed as $\bm{e}_q = \bm{q}_{d,f} - \bm{q}_f$, and 
$\dot{\bm{e}}_q = \dot{\bm{q}}_{d,f} - \dot{\bm{q}}_f$, respectively. The parameters $\bm{K}_q \in \mathbb{R}^{n \times n}$ and $\bm{D}_q \in \mathbb{R}^{n \times n}$ represent the positive definite stiffness and damping matrices for the tracking controller. Moreover, $\bm{q}_{d,f}, \dot{\bm{q}}_{d,f}$ are the position and velocity of the leader that are received on the follower side as desired values.

In order to preserve passivity and thus stability of teleoperation, the Time Domain Passivity Approach (TDPA), adapted from \cite{chen2023passivity}, is implemented in this study. The effect of time delay falls outside the scope of this work and is not considered. 

\section{Method: Shared Tele-assembly}
\label{sharing strategy}


Direct teleoperation heavily centralizes autonomy on the human side. In a conventional bilateral setup, the leader robot functions solely as an interface, faithfully transmitting the operator's commands to the follower robot and rendering sensory feedback to the operator, while the follower robot directly executes these commands to track the operator's trajectory. 

However, executing high-precision, tight-clearance assembly through direct teleoperation introduces severe cognitive and physical challenges. First, camera perspective distortion and viewpoint misalignment significantly impair the operator's spatial perception. Second, precisely regulating a high-dimensional, multi-DoF device demands exceptional motor skills. Consequently, achieving simultaneous and precise control over both the position and orientation of the manipulated object becomes exceedingly difficult, often leading to task failures during contact-rich phases.

To mitigate these limitations, we propose a shared-autonomy strategy that decomposes the manipulation authority into hierarchical subtasks of varying granularity, redistributing the workload across the human operator, the leader, and the follower:
\begin{itemize}
    \item The \textbf{human operator} retains high-level authority, focusing primarily on the \textit{coarse-grained} task of guiding the global trajectory to progress the overall assembly.
    \item The \textbf{leader robot} manages the \textit{medium-grained} task by providing active physical guidance (e.g., via soft virtual fixtures), assisting the operator in maintaining proper orientation alignment by constraining unnecessary control DoFs throughout the task.
    \item The \textbf{follower robot} autonomously handles the \textit{fine-grained} task, executing operator's command and fine-tuning the object's orientation during the final critical insertion stage.
\end{itemize}



\begin{table*}[htbp]
\centering
    \begin{tabular}{cllccccc}
        \toprule
         \multirow{2}{*}{\textbf{Task}} & \multirow{2}{*}{\textbf{Peg Shape}} & \multirow{2}{*}{\textbf{Peg Geometry (mm)}} & \multicolumn{2}{c}{\textbf{Peg}} & \multicolumn{2}{c}{\textbf{Hole}} & \multirow{2}{*}{\textbf{Clearance (mm)}} \\
        \cmidrule(lr){4-5} \cmidrule(lr){6-7}
        & & & \textbf{Material} & \textbf{Manufacturing} & \textbf{Material} & \textbf{Manufacturing} & \\
        \midrule
        A & Cylinder & length: 50, diameter: 40 & PAM & CNC & PLA & 3D printing & 0.25 \\
        B & Cylinder & length: 50, diameter: 40 & PAM & CNC & PLA & 3D printing & 0.13 \\
        C & Cylinder & length: 50, diameter: 40 & PAM & CNC & PLA & 3D printing & 0.07 \\
        D & Cuboid & length: 35, width: 25, height: 60 & PLA & 3D printing & PLA & 3D printing & 0.10 \\
        E & Hexagonal Prism & length: 50, side length: 11 & PLA & 3D printing & PLA & 3D printing & 0.10 \\
        F & Cylinder & length: 50, diameter: 20 & PAM & CNC & Aluminum & CNC & 0.02 \\
        \bottomrule
    \end{tabular}
    \caption{Task objects configuration. Tasks A, B, and C share the same peg, differing only in hole size.}
    \label{tab:objects}
\end{table*}

Building on this autonomy sharing strategy, the original bilateral teleoperation framework is enhanced as follows:

\subsubsection{Shared control on leader side}
To preserve the operator's high-level authority, the proposed framework implements a shared-control paradigm on the leader side that selectively constrains the controllable DoF, thereby alleviating the operator's cognitive burden. Under this scheme, the human operator retains direct, intuitive control over coarse, task-critical dimensions, while the leader robot autonomously stabilizes or constrains complementary, non-critical dimensions according to dynamic task requirements. 



Most sequential assembly procedures, such as insertion, fastening, and press-fitting, are geometrically governed by a prominent assembly axis $\bm{z}_{\text{axis}}$, which defines the feasible motion direction for the mating parts to be joined. Concurrently, the remaining spatial dimensions function as a constrained subspace, within which the components must strictly adhere to rigid geometric boundaries to enable the relative insertion movement along a $\bm{z}_{\text{axis}}$. In practice, simultaneously maintaining these multi-axial geometric boundaries while advancing the parts makes tight-clearance assembly exceptionally difficult, particularly in teleoperation scenarios. To alleviate this operational strain, we implement a shared-autonomy strategy that decouples the task space into parallel control objectives: the operator retains direct authority over the dynamic translation along and rotation around the feasible motion axis assembly axis $\bm{z}_{\text{axis}}$, while the leader robot concurrently regulates assembly axis's orientation to ensure robust orientation alignment.



To align the $z$-axis $\bm{z}_l$ of End-Effector (EE) with the target direction $\bm{z}_{axis}$ via minimal intervention, the leader's desired goal orientation is determined by rotating the current EE frame's orientation by an angle $\theta$ along the axis $\bm{a}_r$, as defined by the following equations:
\begin{equation}
\begin{aligned}
    \bm{a}_{r} &= \bm{z}_l \times \bm{z}_{axis}, \\
    \theta &=  \operatorname{arccos}(\bm{z}_l \cdot \bm{z}_{axis}).
\end{aligned}
\end{equation}
This angular error is subsequently converted into Euler angles to compute the desired goal pose $\bm{p}_d$. The original leader's control law is enhanced as:
\begin{equation}
\begin{aligned}
    \bm{\tau}_{c,l} &= \bm{J}_{l}^{T}( \bm{q}_l) \bm{\Lambda}_1 [\bm{K}_{c} (\bm{p}_d - \bm{p}_{l}) - \bm{D}_c \dot{\bm{p}}_{l}]\\
    &+  \bm{c}_l(\bm{q}_l, \dot{\bm{q}}_l) + \bm{g}(\bm{q}_l) + \bm{\tau}_{d,f}, 
\end{aligned}
\end{equation}
where $\bm{J}_l \in \mathbb{R}^{6 \times 7}$ is the Jacobian matrix of the leader robot, and $\bm{\Lambda}_1 \in \mathbb{R}^{6 \times 6} $  is a selection matrix calculated from the insertion axis\footnote{Without loss of generality, our real-world evaluation configures the assembly axis along the global vertical direction, i.e., $\bm{z}_{\text{target}} = [0, 0, 1]^T$. This specific geometric setup yields $\bm{\Lambda}_1 = \operatorname{diag}(0, 0, 0, 1, 1, 0)$, which isolates off-axis rotational deviations into the constraint-satisfaction subspace.}. Moreover matrices $\bm{K}_c \in \mathbb{R}^{6 \times 6}$ and $\bm{D}_c \in \mathbb{R}^{6 \times 6}$ represent the Cartesian stiffness and damping matrices. The leader's current pose and velocity are expressed as $\bm{p}_{l}$ and $\dot{\bm{p}}_{l}$, respectively.


\subsubsection{Shared autonomy on follower side}
In contrast to the leader robot, which provides continuous assistance to the operator throughout the entire assembly process, the autonomy level of the follower robot is adaptively adjusted during task execution based on both measured interaction forces and the requirements of the assembly task. Owing to the limited transparency inherent in teleoperation systems, directly tracking the leader's motion can lead to prolonged execution times and, in some cases, task failure, particularly in tight-clearance assembly scenarios. To mitigate these limitations, we incorporate force-domain knowledge from robotic assembly into the shared-control framework through a feedforward wrench term, $\bm{f}_{ff} \in \mathbb{R}^{6}$, expressed in the EE frame. The follower control law is therefore modified as: 
\begin{equation}
\begin{aligned}
    \bm{\tau}_{c,f} &= \bm{K}_q \bm{e}_q + \bm{D}_q \dot{\bm{e}}_q + \bm{c}_f(\bm{q}_f, \dot{\bm{q}}_f) + \bm{g}_f(\bm{q}_f) \\ & +  \eta  \bm{J}_{b,f}^{T}( \bm{q}_f) \bm{\Lambda}_2  \bm{f}_{ff}.
\end{aligned}
\end{equation}
where $\bm{J}_{b,f} \in \mathbb{R}^{6 \times 7}$ denotes the body Jacobian of the follower robot, which maps joint velocities to the EE twist expressed in the body (EE) frame. The matrix $\bm{\Lambda}_2 \in \mathbb{R}^{6 \times 6}$ is a diagonal selection matrix that specifies the dimensions along which the assembly force-domain knowledge is applied. The assembly skill is regulated by the autonomy level $\eta$, which is determined by the operator's intent and current execution situation. Let $\mathcal{C}(\bm{f}_{ext,f})$ denote a task-specific condition defined according to the assembly requirements. The autonomy level is then given by
\begin{equation}
    \eta = 
    \begin{cases}
        1, & \mathcal{C}(\bm{f}_{ext,f}) \text{ holds}, \\
        0, & else,
    \end{cases}   
\end{equation}

To increase the legibility and predictability of the system, and reduce the disturbance applied to the operator, the part of the external force which is aligned with $\bm{f}_{ff}$ is not transmitted back to the operator. Therefore, the feedback torque transmitted to the leader $\bm{\tau}_{ext,f}'$ is:
\begin{equation}
\begin{aligned}
    \bm{\tau}_{ext,f}' &= \bm{J}_{b,f}^{T}( \bm{q}_f)(\bm{I}_6 - \bm{\Lambda}_2) \bm{f}_{ext, f}, \\ 
\end{aligned}
\end{equation}
where $\bm{f}_{ext,f} \in \mathbb{R}^{6}$ denotes the external wrench acting on the follower end-effector, expressed in the EE frame.

For the sub-millimeter tight clearance insertion task, the assembly skill is a wiggle motion, which has been shown to improve insertion robustness in tight-clearance assembly scenarios \cite{johannsmeier2019framework, wu20241}. Accordingly, the feedforward term is designed as: 
\begin{equation}
    f_{ff,j}(t) =  a_j \cdot \sin( 2\pi f_jt + \varphi_j), \label{Lissajous} 
\end{equation}
where $a_j$, $f_j$ and $\varphi_j$ refer to the amplitude, frequency and phase, respectively. The subscript $j$ refers to the direction in the range $\{x_f, y_f, z_f, rx_f, ry_f, rz_f\}$ of the EE frame, and $t$ indicates the elapsed execution time. 

According to the proposed autonomy-sharing mechanism, the robot assumes control over the Cartesian pose's $rx$ and $ry$ components, preventing the user from experiencing external forces along the corresponding axes. Consequently, the selection matrix $\bm{\Lambda}_2 = \operatorname{diag}(0, 0, 0, 1, 1, 0)$ is chosen to apply the force domain knowledge on the corresponding dimension. The assembly skill is activated when the insertion process generates sufficiently large contact forces, thus the condition $\mathcal{C}(\bm{f}_{ext,f})$  is defined as $\parallel \bm{f}_{ext,f,xy} \parallel > f_t$, where $\bm{f}_{ext,f,xy}$ denotes the component of the measured interaction force in the $xy$ plane of the EE frame, and $f_t$ is a predefined force threshold.

\section{Experiment}

To evaluate the performance of the proposed \textit{SharedAssembly} on sub-millimeter-clearance assembly tasks, we design the subsequent experiments to address the following research questions:
\begin{itemize}
    \item \textbf{RQ1}: How does the proposed method perform compared with existing teleoperation methods? 
    \item \textbf{RQ2}: Can it narrow the performance gap between novice and experienced operators? 
\end{itemize}

\subsection{Experimental Setup}

\begin{figure}[h]
    \centering
    \includegraphics[width=0.95\linewidth]{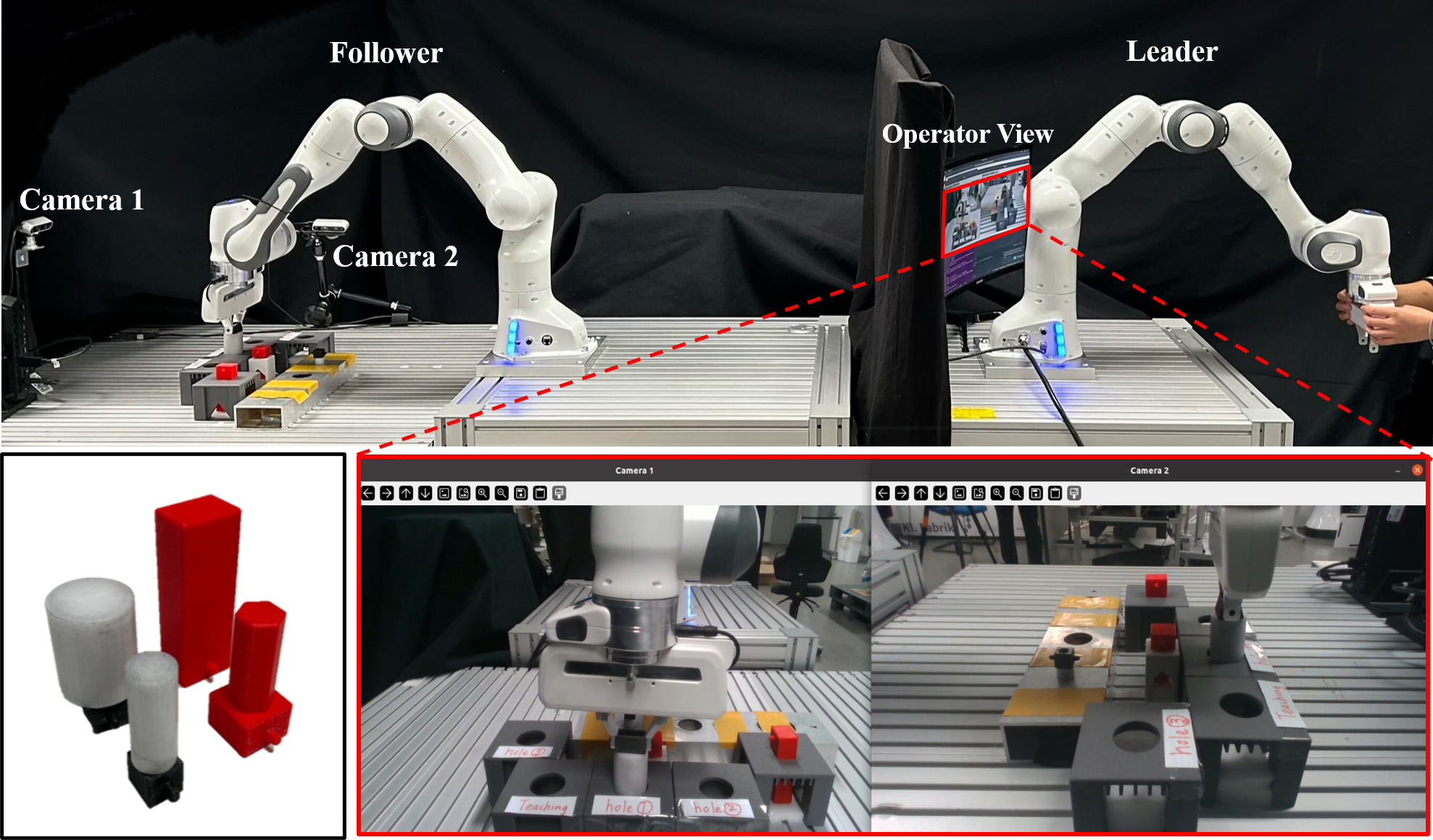}
    \caption{Experimental setup for data collection based on shared tele-assembly}
    \vspace{8pt}
    \label{fig:setup}
\end{figure}

As illustrated in Fig.~\ref{fig:setup}, the experimental setup comprises three platforms (arranged from left to right) to accommodate the insertion tasks, the follower robot, and the leader robot, respectively. The detailed configuration of each insertion object is provided in Table~\ref{tab:objects}. Both manipulators are Franka Emika Panda robots. The leader robot is controlled by a PC running Ubuntu 20.04 with Intel Core i7-10700 CPU, while the follower is controlled by another PC running Ubuntu 20.04 with Intel Core i9-11900K CPU. Two PCs communicate through User Datagram Protocol (UDP) within the same network, making network delay negligible. Additionally, two Intel RealSense D435i cameras are mounted on the follower’s side, positioned vertically to provide stereo views for the human operator.

\subsection{Compared Methods \label{sec:baseline}}

To evaluate the effectiveness of the proposed shared-autonomy framework, we compare it with representative teleoperation methods with different levels of feedback and autonomy: 
\begin{itemize}
    \item \textbf{Unilateral Teleoperation: }A conventional and most widely used teleoperation framework, where  the operator controls the follower robot solely based on visual information. 
    \item  \textbf{Bilateral Teleoperation}:  A teleoperation framework that enables the operator to perceive interaction forces occurring at the follower side through haptic feedback on the leader side~\cite{chen2024network}.
    \item \textbf{Shared Bilateral Teleoperation}: The proposed framework that integrates bilateral force feedback and shared autonomy.
\end{itemize}

\subsection{Experimental Procedure \label{sec:experience-level}}
To ensure reliable results, 30 participants were recruited for the user study. Due to invalid experimental data from one participant, 29 participants were included in the final analysis. Based on their prior experience with teleoperation and robotics, the participants were categorized into three groups: 
\begin{itemize}
    \item \textbf{Expert} ($n=8$): Participants with teleoperation experience.
    \item \textbf{Intermediate} ($n=10$): Participants with prior experience in robotics but not teleoperation.
    \item \textbf{Novice} ($n=11$): Participants with no prior experience in robotics.
\end{itemize}

The user study procedure consists of three phases: \textbf{Phase I (Training):} Participants complete three training trials using a hole with a relatively large clearance to familiarize themselves with the teleoperation system and the assembly task. \textbf{Phase II (Clearance Evaluation):} Tasks A, B, and C are performed to evaluate the influence of assembly clearance on teleoperation performance. \textbf{Phase III (Shape Evaluation):} Tasks D, E, and F are performed to investigate the impact of different peg geometries on task performance.

For each task, every participant performed three repetitions under each of the three teleoperation methods. To minimize the effects of \textit{sensorimotor adaptation} during skill acquisition~\cite{burdet2013human}, both the task order and teleoperation method were randomized across trials. Furthermore, participants were blinded to the teleoperation method being evaluated in order to reduce potential bias arising from cognitive expectations.

\subsection{Evaluation Metrics}

\subsubsection{Success Rate}Success rate is defined as the percentage of successful trials among all conducted trials:

\begin{equation}
SR = \frac{N_{\mathrm{success}}}{N_{\mathrm{total}}},
\end{equation}

where $N_{\mathrm{success}}$ denotes the number of successful trials and $N_{\mathrm{total}}$ denotes the total number of trials. A trial is considered unsuccessful if it fails to complete the assembly task within $60$~s or violates the robot safety constraints. 

\subsubsection{Task Completion Time}
To provide a more detailed analysis, the tele-assembly process is divided into two stages: (i) \textit{Move-to-Contact}, in which the operator moves the peg from its initial position to the vicinity of the hole, and (ii) \textit{Guided Insertion}, in which the peg is inserted into the hole until task completion. The corresponding completion times of the two stages in the $i$-th trial are denoted by $t_{\mathrm{move},i}$ and $t_{\mathrm{insert},i}$, respectively. 

Based on this decomposition, two temporal metrics are evaluated, i.e., the \textit{average guided insertion time} $\bar{t}_{\mathrm{insert}}$ (Stage II only) :
\begin{equation}
\begin{aligned}
    \bar{t}_{\mathrm{insert}} &= \frac{1}{N_{\mathrm{total}}}
    \sum_{i=1}^{N_{\mathrm{total}}} t_{\mathrm{insert},i},\\
\end{aligned}
\end{equation}
and the \textit{average entire assembly time} $\bar{t}_{\mathrm{entire}}$ (the combined duration of both stages):
\begin{equation}
\begin{aligned}
    \bar{t}_{\mathrm{entire}} &= \frac{1}{N_{\mathrm{total}}}
    \sum_{i=1}^{N_{\mathrm{total}}} t_{\mathrm{entire},i} ,
\end{aligned}
\end{equation}
where $t_{\mathrm{entire},i}$ is assigned a value of $60$~s for failed trials: 
\begin{equation}
    t_{\mathrm{entire},i} =\begin{cases}
  t_{\mathrm{move},i} + t_{\mathrm{insert},i}, & \text{successful trial},\\
60, & \text{failed trial}.
\end{cases}\\
\end{equation}
\subsubsection{Data Collection Efficiency}
Following the Common Industry Format for Usability Test Reports (\texttt{ISO/IEC 25062:2006})~\cite{albert2013measuring}, we use the efficiency metric $\eta_c$ to measure the method's data collection efficiency, defined as the ratio of the task completion rate to the mean time per task (measured in minutes):
\begin{equation}
    \eta_c = \frac{SR}{\bar{t}_{\mathrm{entire}}} \times 60.
\end{equation}

\section{Results and Discussion}
\subsection{RQ1: Overall Performance Comparison}
To answer this research question, we compare the proposed method with existing teleoperation approaches (see Sec.~\ref{sec:baseline}). The results show that our SharedAssembly delivers the strongest overall performance in terms of both success rate and task completion time. 

\begin{table}[htbp]
\centering
\begin{tabular}{l@{\hskip 7pt}c@{\hskip 7pt}c@{\hskip 7pt}c@{\hskip 7pt}c@{\hskip 7pt}c@{\hskip 7pt}c@{\hskip 7pt}c}
\toprule
\textbf{Method} & \textbf{A} & \textbf{B} & \textbf{ C} & \textbf{D} & \textbf{E} &  \textbf{F} & \textbf{Average}\\ 
\midrule
Unilateral & 96.3  &  75.3& 48.2& 63.2& 73.3
&  48.1& 67.4\\
Bilateral  &  \textbf{100.0} &  95.3& 66.7&  94.7& 
\textbf{95.1}& 90.4& 91.0\\
SharedAssembly & \textbf{100.0}  & \textbf{100.0}  &  \textbf{94.1}&  \textbf{100.0} & 
93.9& \textbf{94.1}& \textbf{97.0}\\
\bottomrule
\end{tabular}
\caption{Success Rate [\%]}
\label{tab:succ}
\end{table}

The success rates for all evaluated cases are summarized in Table~\ref{tab:succ}. Overall, teleoperation methods with force feedback (SharedAssembly and Bilateral) constantly achieve substantially higher success rates than unilateral teleoperation, highlighting the importance of haptic feedback in tight clearance tele-assembly tasks. Additionally, compared with bilateral teleoperation, our SharedAssembly achieves superior performance in most experimental conditions, further increasing the success rate from 91\% to 97\%. This result indicates that the proposed shared-autonomy strategy provides additional benefits beyond force feedback alone.

\begin{figure}[]
    \centering
    \includegraphics[width=0.98\linewidth]{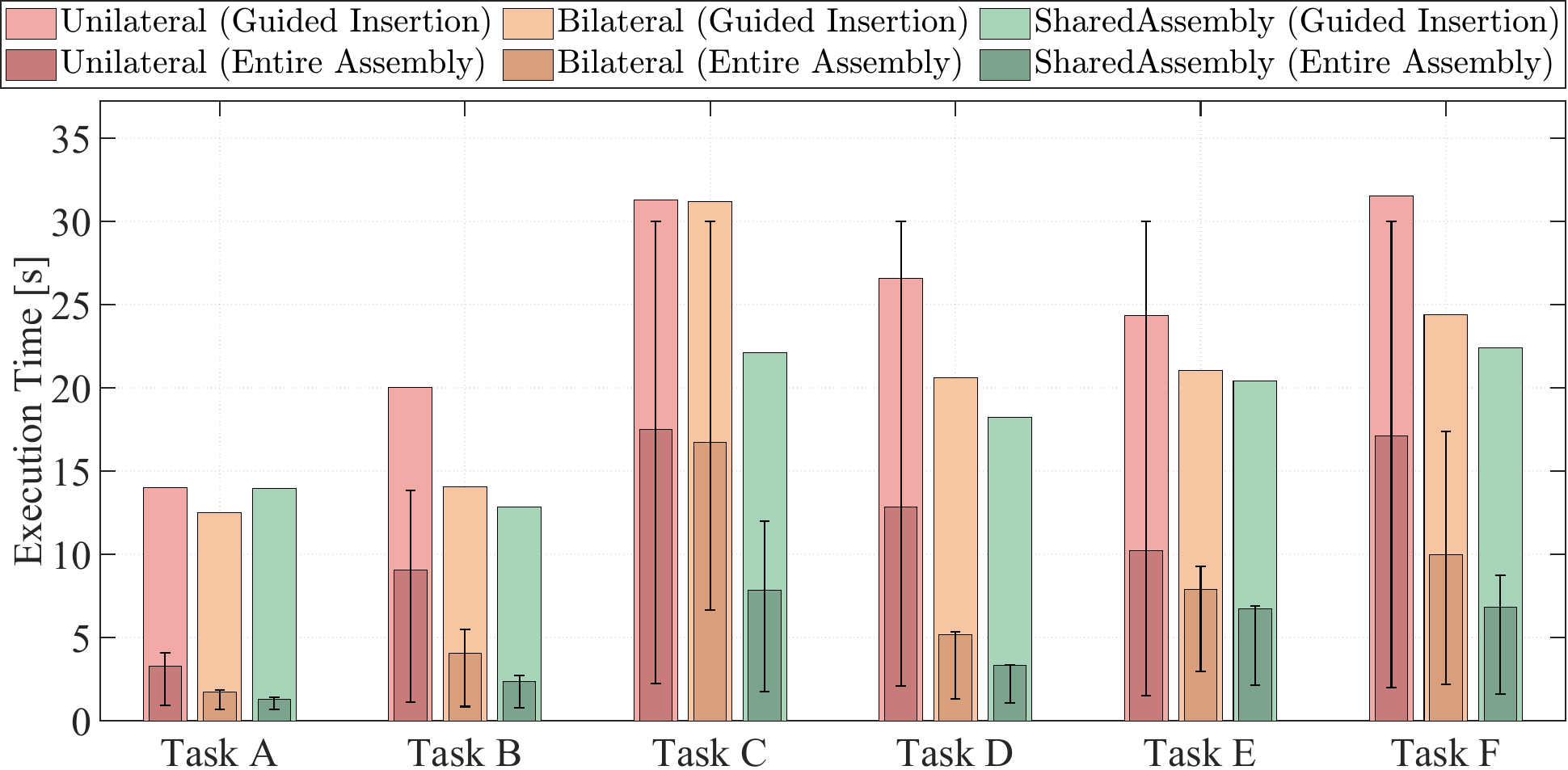}
    \caption{Task execution time. The darker-color columns with interquartile range indicators represent the execution time during the guided insertion phase; The lighter-color columns correspond to the total execution time, encompassing both the position guiding phase and the guided insertion phase.}
    \label{fig:result1}
\end{figure}

The task completion time shown in Fig.~\ref{fig:result1} exhibits a similar trend. When focusing on the guided insertion stage, where contact interactions primarily occur, the corresponding average insertion time $\bar{t}_{\mathrm{insert}}$ decreases monotonically from unilateral to bilateral teleoperation and further to SharedAssembly. Meanwhile, our approach exhibits the highest robustness, as indicated by the narrowest interquartile range. This result indicates that the proposed shared-autonomy framework is particularly effective in assisting the contact-rich insertion process, thereby reducing the time required for the fine corrective motions that are essential in tight-clearance tele-assembly tasks. 

Moreover, the benefits of force feedback and shared autonomy become increasingly pronounced as task difficulty increases. Tasks A, B, and C share the same peg geometry but differ in assembly clearance (see Table~\ref{tab:objects}), resulting in progressively more challenging assembly conditions. Correspondingly, the success rates of unilateral and bilateral teleoperation drop by approximately 50\% and 33\%, respectively. In contrast, SharedAssembly maintains a success rate above 94\% even under the tightest clearance condition. Meanwhile, the reduction in guided insertion time achieved by the proposed method becomes increasingly significant, further demonstrating its effectiveness in addressing tight-clearance tele-assembly tasks.

\subsection{RQ2: Performance Across Different Experience Levels}
To address this question, we further analyze the experimental results by grouping participants according to their teleoperation experience levels, as described in Sec.~\ref{sec:experience-level}. While operator experience has a significant impact on tele-assembly performance in terms of success rate and data collection efficiency, the proposed SharedAssembly substantially reduces this dependence across all experience groups.

\begin{figure}[h]
    \centering
    \vspace{-3pt}
    \includegraphics[width=0.95\linewidth]{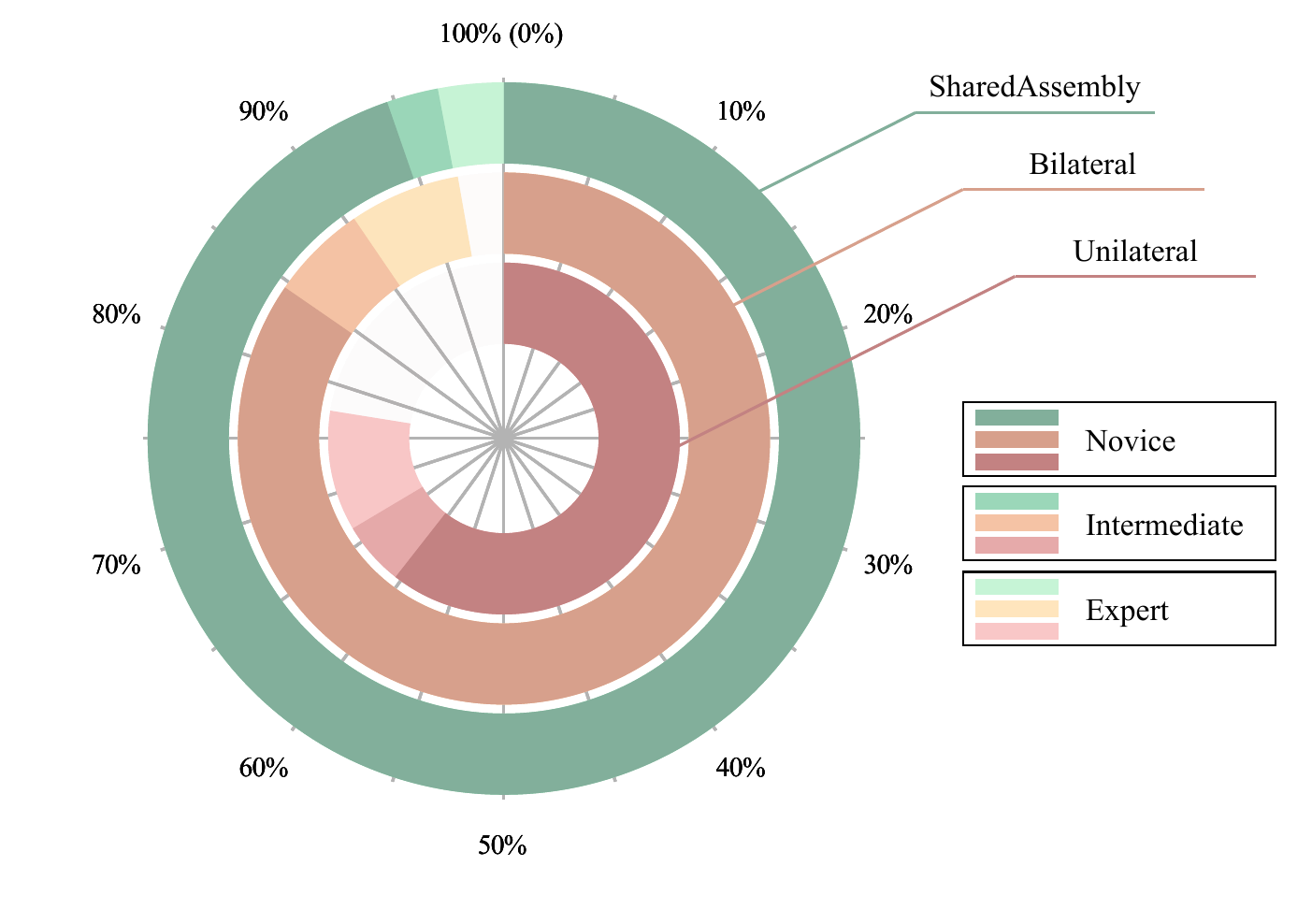}
    \vspace{-11pt}
    \caption{Overall success rate for all tasks across participants with different experience levels.}
    \vspace{6pt}
    \label{fig:result2}
\end{figure}

\begin{figure}[t]
    \centering
    \includegraphics[width=0.98\linewidth]{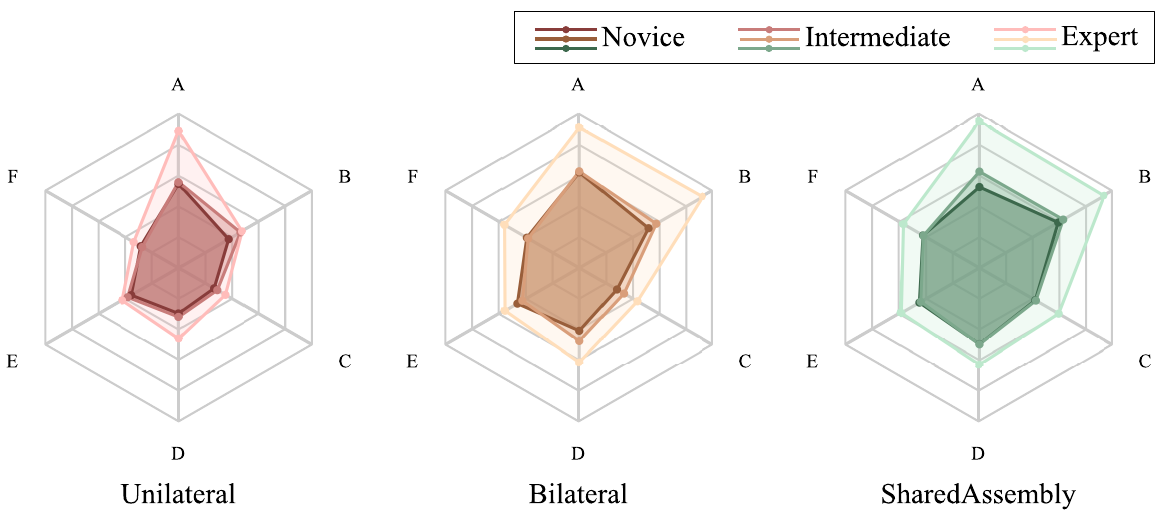}
    \caption{Data collection efficiency across participants with different experience levels.}
    \label{fig:result3}
\end{figure}

Fig.~\ref{fig:result2} depicts the tele-assembly success rates achieved by participants with different levels of teleoperation experience, while Fig.~\ref{fig:result3} illustrates the corresponding data collection efficiency. As illustrated in the unilateral teleoperation results, a substantial performance gap exists between novice and expert operators in both metrics. This finding highlights the strong reliance of teleoperation-based data collection on experienced operators, whose expertise typically requires extensive training and practice to develop.

More importantly, the proposed method substantially narrows the performance gap between operators with different experience levels, enabling less experienced operators to achieve performance comparable to that of more experienced users. As illustrated in Fig.~\ref{fig:result2}, intermediate operators assisted by SharedAssembly even outperform expert operators using bilateral teleoperation, achieving a success rate of 97.2\%. Furthermore, novice and intermediate operators using SharedAssembly achieve performance comparable to that of expert operators using the baseline methods. These findings indicate that successful tele-assembly becomes significantly less dependent on operator expertise, thereby enabling more scalable and accessible demonstration collection for contact-rich manipulation tasks.

Notably, the benefits of SharedAssembly are not limited to less experienced operators. As shown in Fig.~\ref{fig:result2} and Fig.~\ref{fig:result3}, substantial performance improvements can be observed across all experience groups. While the largest gains are achieved by novice and intermediate operators, expert users also benefit from the proposed framework in terms of both success rate and task efficiency. Consequently, the proposed framework demonstrates strong potential to facilitate rapid and large-scale teleoperation-based data collection for contact-rich manipulation tasks.

\section{CONCLUSION}
In this work, we present a novel shared-autonomy-based Bilateral Teleoperation approach for tight-clearance assembly tasks. By integrating task-specific assembly knowledge into both the leader and follower sides of the teleoperation system, the proposed framework assists operators during the contact-rich insertion process while preserving human authority.

Experimental evaluations on \textit{sub-millimeter-clearance} assembly tasks demonstrated the effectiveness of the proposed approach. Compared with conventional unilateral and bilateral teleoperation methods, SharedAssembly achieved superior overall performance in terms of assembly success rate and task completion efficiency. Furthermore, its advantages became increasingly pronounced as assembly clearance decreased and task difficulty increased, highlighting the robustness of the proposed shared-autonomy strategy in demanding contact-rich manipulation scenarios. Beyond improving task performance, the proposed framework substantially reduced the dependence of successful tele-assembly on operator expertise, enabling less experienced users to achieve performance comparable to that of expert operators using conventional teleoperation methods. Meanwhile, experienced expert operators also benefited from the proposed framework, demonstrating the broad applicability of the shared-autonomy strategy across operators with different levels of expertise. As a result, it enables more scalable and accessible demonstration collection for contact-rich manipulation tasks, providing a practical pathway toward large-scale data acquisition for robot learning.

Future work will investigate adaptive autonomy allocation strategies and extend the proposed framework to a broader range of contact-rich manipulation tasks beyond peg-in-hole assembly, further exploring its potential for general-purpose robotic manipulation and robot learning applications.


\section*{ACKNOWLEDGMENT}
We sincerely thank the 30 volunteers for their participation in our experiment and their valuable contributions to our data collection as well as Prof. Alexander König for his valuable comments. The authors would like to thank the Federal Ministry of Research, Technology, and Space (BMFTR) for its support as part of the research program Communication Systems “Souverän. Digital. Vernetzt.”. Joint project 6G-life, project identification number: 16KIS2414. We gratefully also acknowledge the funding of the Lighthouse Initiative Geriatronics by LongLeif GaPa gGmbH (Project Y) and the funding of the Lighthouse Initiative KI.FABRIK Bayern by StMWi Bayern, Forschungs- und Entwicklungsprojekt, Grant No. DIK0249.


\bibliography{IEEEabrv,mybib2022}
\bibliographystyle{myIEEEtran}


\end{document}